\documentclass[letterpaper]{article} 
\usepackage{aaai25}  
\usepackage{times}  
\usepackage{helvet}  
\usepackage{courier}  
\usepackage[hyphens]{url}  
\usepackage{graphicx} 
\usepackage{natbib}  
\usepackage{caption} 
\usepackage{algorithm}
\usepackage{algorithmic}

\usepackage{newfloat}
\usepackage{listings}
\DeclareCaptionStyle{ruled}{labelfont=normalfont,labelsep=colon,strut=off} 
\floatstyle{ruled}
\newfloat{listing}{tb}{lst}{}
\floatname{listing}{Listing}
\title{M4CXR: Exploring Multi-task Potentials of Multi-modal Large Language Models for Chest X-ray Interpretation}
\author{
    Jonggwon Park, 
    Soobum Kim, 
    Byungmu Yoon, 
    Jihun Hyun,
    Kyoyun Choi\thanks{Corresponding author}
    \nocopyright
}
\affiliations{
    DEEPNOID Inc., Seoul, Korea\\
    \{jgpark, soobumk, bmyoon, jhyun, kychoi\}@deepnoid.com
}

\usepackage{bibentry}

\usepackage{kotex}
\usepackage{nicematrix} 
\usepackage{multirow} 
\usepackage{booktabs} 
\usepackage{arydshln} 
\usepackage{amsmath} 
\usepackage{xcolor}
\usepackage{hhline}
\usepackage{pifont}
\usepackage{makecell}

\begin{document}

\maketitle
\renewcommand{\arraystretch}{1}

\begin{abstract}
The rapid evolution of artificial intelligence, especially in large language models (LLMs), has 
significantly impacted various domains, including healthcare. 
In chest X-ray (CXR) analysis, previous studies have employed LLMs, but with limitations: either underutilizing the multi-tasking capabilities of LLMs or lacking clinical accuracy.
This paper presents M4CXR, a multi-modal LLM
designed to enhance
CXR interpretation.
The model is trained on a visual instruction-following dataset that integrates various task-specific datasets in a conversational format. 
As a result,
the model supports
multiple tasks
such as medical report generation (MRG), visual grounding, and visual question answering (VQA). 
M4CXR achieves state-of-the-art clinical accuracy in MRG by employing
a chain-of-thought prompting strategy, in which it identifies findings in CXR images and subsequently generates corresponding reports.
The model is adaptable to
various MRG scenarios depending on the available inputs, such as single-image, multi-image, and multi-study contexts.
In addition to MRG, 
M4CXR performs visual grounding 
at a level comparable to
specialized models
and also demonstrates outstanding performance in VQA.
Both quantitative and qualitative assessments reveal M4CXR’s versatility in MRG, visual grounding, and VQA, while consistently maintaining clinical accuracy.

\end{abstract}

\section{Introduction}
Recent advancements in artificial intelligence, particularly large language models (LLMs), have led to their widespread application across various fields,
including healthcare.
Numerous studies are 
exploring diverse methods of application for improving healthcare outcomes, 
such as
personalized treatment plans, clinical decision support systems, and enhancing medical education \cite{medllm-survey}.

Within the medical domain, 
this paper focuses on chest X-ray (CXR) interpretation.
Although many studies have investigated LLMs for CXR, most encounter one of two issues:
they either underutilize the capabilities of LLMs or 
struggle with
ensuring clinical accuracy.
Despite LLMs' ability to perform various tasks through conversational interaction, many studies focus on a single task, 
typically
medical report generation (MRG) 
\cite{lmrrg, maira1, llava-rad}.
While some studies effectively leverage LLMs to 
comprehend questions and generate appropriate responses
\cite{xraygpt, radialog, llmcxr, chexagent},
the clinical accuracy of these conversational outcomes, including generated reports, falls short of expectations.

\begin{figure*}[t]
\centering
\includegraphics[width=1\textwidth]{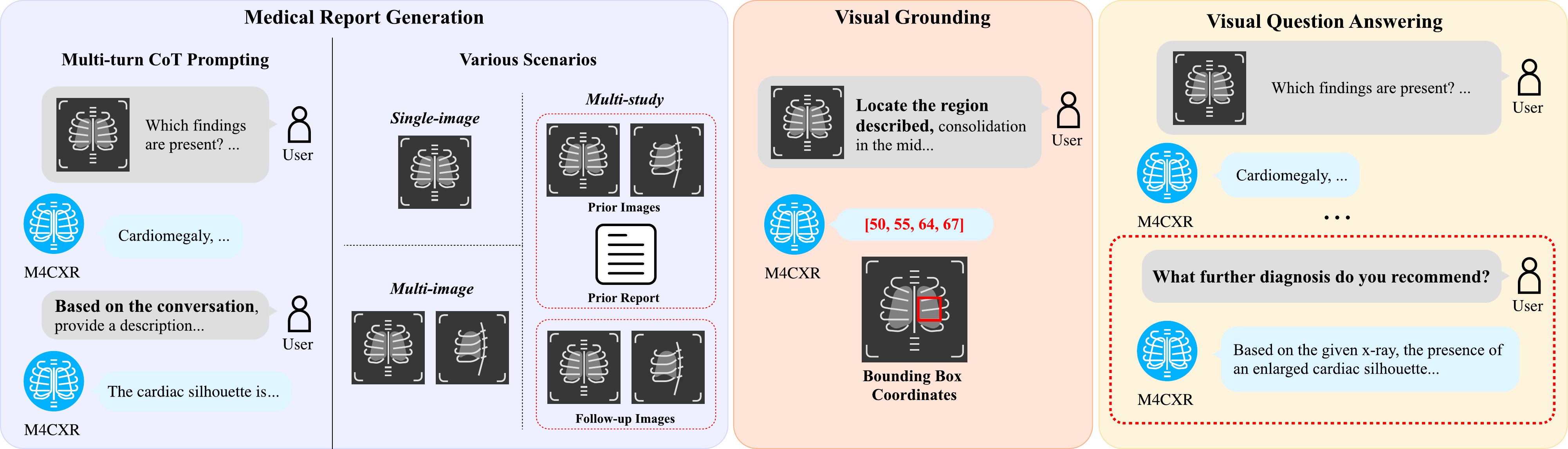} 
\caption{
Overview of the multi-tasking capabilities of M4CXR. 
Facilitated by 
CoT prompting in MRG, M4CXR produces 
clinically accurate reports and adapts to various scenarios. 
Additionally, M4CXR can ground the locations described in the report or 
answer questions
based on chest X-ray images.
}
\label{fig-intro}
\end{figure*}

In this work, we propose 
M4CXR, an LLM for \textbf{CXR} 
adept at
handling four ``Multi'' aspects: \textbf{M}ulti-modal, \textbf{M}ulti-task, \textbf{M}ulti-image input, and \textbf{M}ulti-turn chatting.
We train the model on a dataset constructed by integrating various task-specific datasets,
enabling M4CXR to excel in
MRG, image understanding, and visual question answering (VQA)
tasks,
as depicted in Figure \ref{fig-intro}.
In MRG, M4CXR improves clinical accuracy by leveraging LLMs' reasoning abilities through a chain-of-thought (CoT) \cite{cot} reasoning, where 
the model identifies findings in the image and generates reports based on these results.
Additionally, 
we conduct experiments 
across various scenarios 
with inputs extending beyond a single image,
including
CXR images from different views 
and prior patient studies. 
Besides 
MRG,
M4CXR demonstrates strong multi-tasking capability 
with successful application in visual grounding and VQA, 
showcasing its 
adaptability
in diverse clinical contexts.

In summary, our main contributions are as follows:
\begin{itemize}
\item We propose M4CXR, an MLLM designed for CXR interpretation, capable of handling multiple tasks. 
To enable multi-tasking, we assemble 
a visual instruction-following
dataset from diverse CXR tasks.
\item By adopting a novel CoT reasoning process, M4CXR achieves state-of-the-art clinical accuracy in CXR report generation.
\item M4CXR effectively utilizes multiple images and reports, allowing for its applicability across different scenarios.
\item Beyond MRG, M4CXR demonstrates remarkable performance in visual grounding and VQA.
\end{itemize}

\section{Related Works}
\subsection{Multi-modal Large Language Models}
As research on LLMs progresses, their advantages continue to emerge.
It is widely known that LLMs are capable of multi-task learning \cite{t5}, 
and \citet{cot} showed that LLMs can reason through multi-turn chats using 
CoT
prompting. 
The rise of multi-modal LLMs, 
which utilize the long context length of LLMs to process visual content, 
further amplifies these advantages. 
MLLMs 
can handle tasks that require 
comprehension of spatial regions, and
visual instruction tuning \cite{llava} allows them to answer free-form questions about images.

For MLLMs, bridging the embedding spaces of vision and language 
is 
crucial.
A common
approach is to
freeze the pre-trained vision model and LLM, and train only the bridging module \cite{flamingo, minigpt4}, also known as the projector.
The projector's structure varies,
from a linear layer to Q-Former \cite{blip2}, C-Abstractor \cite{honeybee}, and more.
Drawing insights from these works,
we employ a two-phase training strategy: pre-training the projector first,
and then
fine-tuning the entire model with visual instruction tuning.

\subsection{Chest X-ray Report Generation}
CXR report generation via deep learning 
has been extensively studied over time
\cite{cxrrnn, automaticmrg, wrrgmkg, rgrg}.
Since the advent of LLMs, 
training strategies
such as freezing pre-trained models \cite{xraygpt} or
visual instruction tuning \cite{llmcxr}
have been equivalently implemented in the CXR domain.
However, a significant drawback of these models
is the lack of clinical accuracy in the generated reports.
Efforts
to address these shortcomings include 
adopting a separate disease classification module \cite{promptmrg, lmrrg}, 
increasing the visual encoder's resolution
\cite{llava-rad}, 
or harnessing additional input
 \cite{maira2}.
In this work, we take a novel yet simple approach: leveraging the reasoning abilities of LLMs.

\subsection{Multi-Tasking in Chest X-ray Interpretation}
Confining the use of LLMs 
solely to report generation 
underutilizes their potential.
MLLMs that capitalize on the characteristics of LLMs to enable conversations based on CXR images include XrayGPT \cite{xraygpt}, RaDialog \cite{radialog}, and LLM-CXR \cite{llmcxr}.
CheXagent \cite{chexagent} is a multi-tasking CXR 
foundation model trained on various tasks, similar to our approach.
Yet, the clinical accuracy of 
these
models' responses remains suboptimal.

Medical foundation models such as Med-Gemini \cite{medgemini2d} and MedPaLM-M \cite{medpalm-m} 
support CXR interpretation with conversational capabilities,
showing satisfactory
clinical accuracy.
However, these models 
are not trained on 
spatial cognition tasks like visual grounding,
which
involves identifying regions in an image
corresponding to a phrase in the text.
For CXR, 
\citet{MSCXR-ECCV} 
released the benchmark dataset MS-CXR for phrase grounding, and
MedRPG \cite{medrpg} specializes in identifying bounding boxes for phrases in CXR images.
The latest concurrent work by \citet{maira2} introduced MAIRA-2, 
an MRG model for CXR 
capable of grounding but limited to grounded report generation.
In contrast, M4CXR handles multiple tasks such as MRG, VQA, and visual grounding, while maintaining clinical accuracy.

\section{Methods}

\begin{figure*}[t]
\centering
\includegraphics[width=1\textwidth]{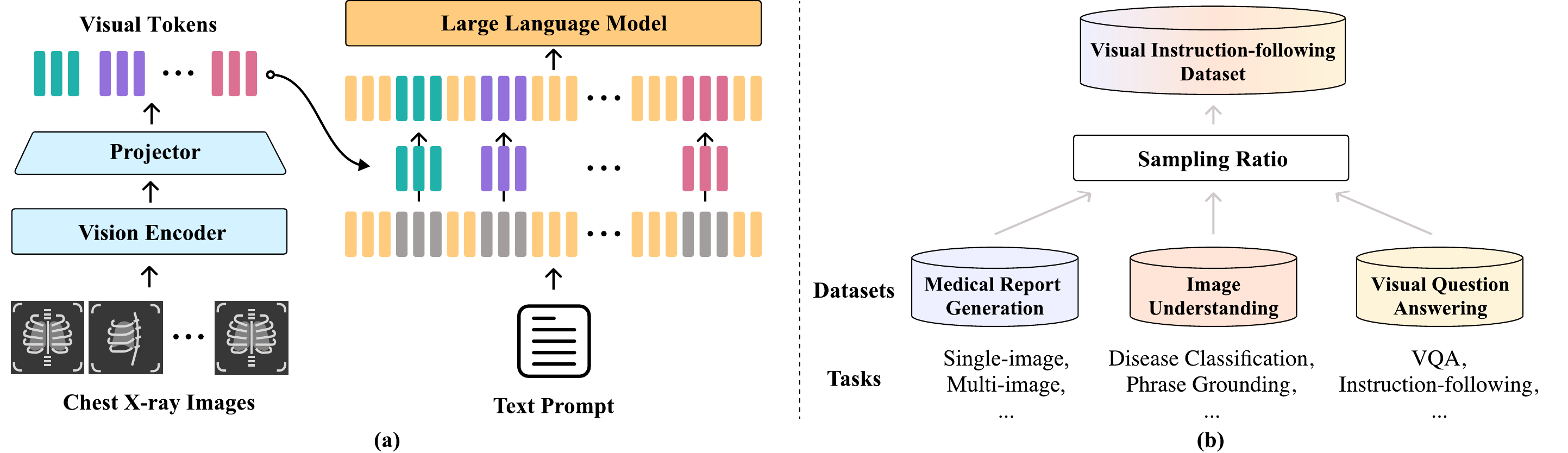} 
\caption{
(a) The architecture of M4CXR.
Utilizing the LLaVA framework, it allows 
visual tokens from each image to be inserted
at designated positions among the text tokens. 
(b) Schema for constructing a CXR visual instruction-following dataset. 
Diverse tasks of three types are combined with appropriate sampling ratios.
}
\label{fig-method}
\end{figure*}

\subsection{M4CXR}
\subsubsection{Architecture.}

Figure \ref{fig-method} (a) 
illustrates the overall architecture of M4CXR.
Following
LLaVA \cite{llava}, M4CXR 
includes a vision encoder, a projector, and an LLM,  
denoted as \(\mathcal{M}\)$_{V}$, \(\mathcal{M}\)$_{P}$, and \(\mathcal{M}\)$_{L}$, respectively. 
Given 
\(n\) CXR images \(x_1, \ldots, x_n\), 
the vision encoder converts each image \(x_i\) into visual feature \(f^v_i\),
which the projector then transforms
into a sequence of visual token embeddings 
\( X^v_i \):
\begin{equation}
X^v_i = \mathcal{M}_{P} (f^v_i) = \mathcal{M}_{P} (\mathcal{M}_{V}(x_i)), \quad \text{for } i = 1, \ldots, n
\end{equation}
A text prompt
is also mapped into the 
embedding space,
resulting in a sequence of language token embeddings \(X^l\).
The LLM takes an input in which visual embeddings are interleaved with language embeddings to generate output \(Y\):
\begin{equation}
Y = \mathcal{M}_{L} ( X^v_1, \ldots, X^v_n, X^l)
\end{equation}
The output of the MLLM \( Y = \{ y_t \}_{t=1}^{T} \),
consisting of \( T \) language tokens, 
is generated 
auto-regressively:
\begin{multline}
p(Y | X^v_1, \ldots, X^v_n, X^l) = \\
\prod_{t=1}^{T} p(y_t | X^v_1, \ldots, X^v_n, X^l, y_{<t}).
\end{multline}

\subsubsection{Bounding Box Representation.}
To enable visual grounding, MLLMs need to represent spatial information as text tokens. 
Referring to \citet{ferret}, we use bounding box coordinates enclosed in square brackets without any additional special tokens. 
The coordinates [\(x_1\), \(y_1\), \(x_2\), \(y_2\)] represent the top-left (\(x_1\), \(y_1\)) and bottom-right (\(x_2\), \(y_2\)) points of the bounding box on the image.
As input images are preprocessed to a uniform size by the vision encoder, the bounding box coordinates are also normalized \cite{minigptv2} to integer values 
between 0 and 100.

\subsection{Multi-turn Chain-of-Thought Prompting} 

\begin{figure}[t]
\centering
\includegraphics[width=\columnwidth]{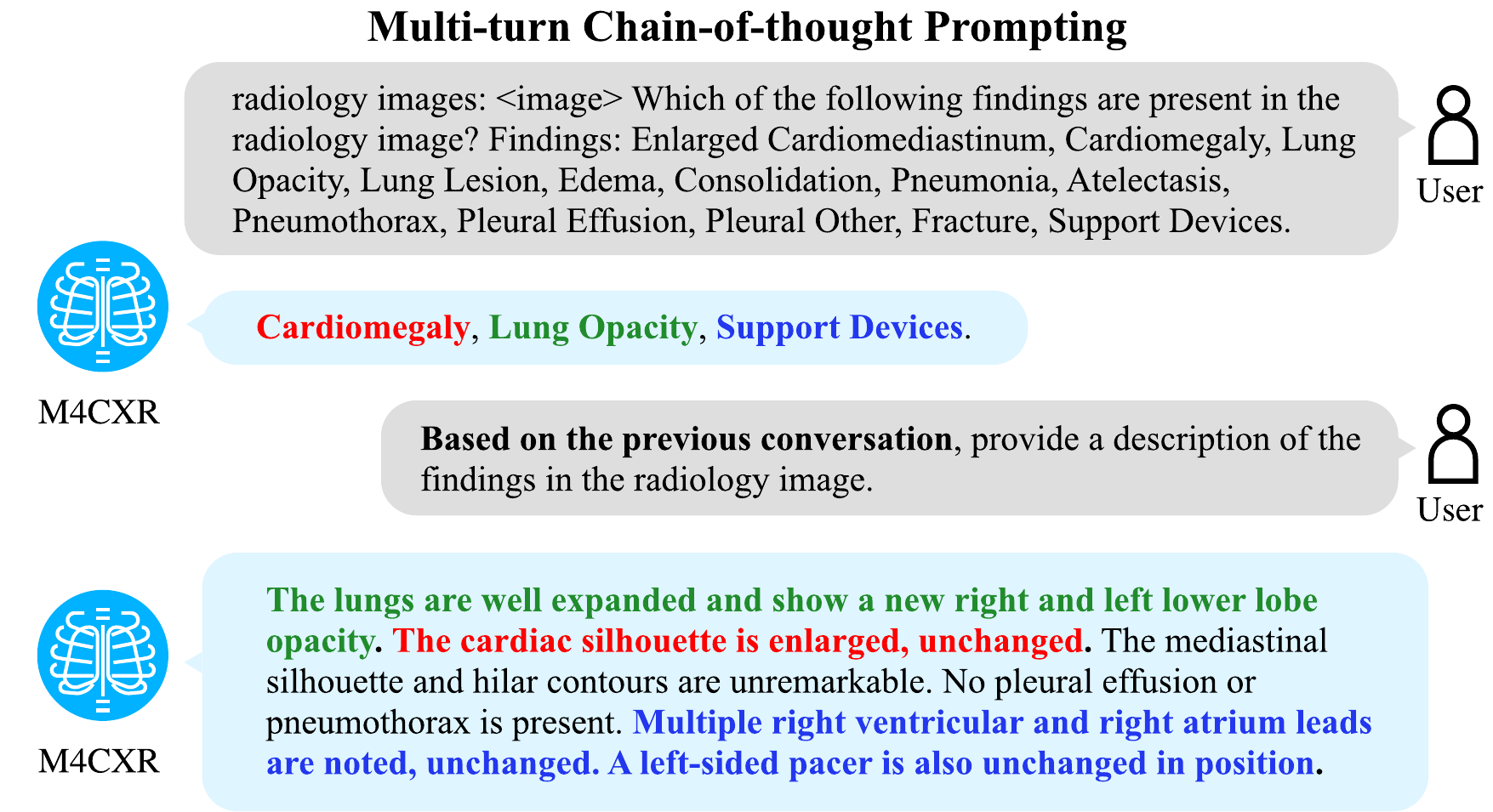} 
\caption{
Example of multi-turn CoT prompting. 
M4CXR first identifies findings in the CXR image, then generates a report. 
Findings and corresponding sentences are color-matched for readability.
}
\label{cot}
\end{figure}

We derive insights from PromptMRG \cite{promptmrg} to enhance the clinical accuracy of MRG.
It classifies observable lesions from a CXR image and then uses the result as input prompts to generate reports.
We follow a similar two-step process,
first identifying key observations and then generating reports.
The difference from PromptMRG is that, in our work, a single MLLM performs both classification and report generation sequentially, 
enabling end-to-end learning.

The gist is to divide the input prompt into two questions, creating 
multi-turn conversational data.
In the first question, we present a list of potential observations and prompt the model to identify those visible in the CXR image. 
Then, we ask the model to generate a report based on the prior conversation.
This approach resembles the reasoning process of a human radiologist, who first identifies lesions visible in the image and then generates a diagnostic report.
It can be seen as a variant of the commonly used CoT \cite{cot} prompting, hence we call it CoT MRG.
Figure \ref{cot} shows the multi-turn CoT prompting process.
To validate the effectiveness of CoT MRG, we conduct a comparative experiment 
with classification and MRG as separate tasks.

\subsection{Integrating Chest X-ray Interpretation Tasks}

We focus on the following features of an MLLM: 
the capability for reasoning through multi-turn chat, 
the ability to utilize multiple images due to its long context length, 
and the flexibility for multi-task learning.
To exploit these features,
we construct a CXR visual instruction-following dataset,
as schematized in Figure \ref{fig-method} (b).
We define a task set across three task types: MRG, image understanding, and VQA,
and transform the corresponding datasets into conversational data.
Detailed task descriptions and conversation templates are provided in the Appendix (Table \ref{tab:dataset-description}, \ref{tab:dataset-tem}).

\subsubsection{Report Generation: Various Scenarios.}
A single radiologic study can include multiple CXR images taken from different views: posterior-anterior (PA), anterior-posterior (AP), or lateral. 
Additionally, when the same patient undergoes follow-up studies, information from prior studies can also be utilized in the radiologic interpretation.

In contrast to previous studies that generate reports from a single image \cite{promptmrg, llava-rad, maira1, rgrg}, we consider MRG tasks in three different scenarios 
based on
the available inputs.
The \textit{single-image} scenario incorporates one image from a single study.
The \textit{multi-image} scenario accesses multiple images from different views within a single study.
The \textit{multi-study} scenario accepts images and the corresponding report from prior study as inputs, along with follow-up study images.
These scenarios have different conversation templates. 
In this work, since we distinguish tasks by templates, they are treated as separate tasks.
Through these different scenario tasks, we expect the model to understand and fully utilize the given information, enabling it to generate more accurate reports when additional inputs are available.

\subsubsection{Image Understanding.}

Disease classification is one of the core tasks in CXR image understanding. 
We use class label data from various classification datasets.
Since each dataset has its own set of disease labels, we specify the relevant labels for each question in the prompt.
The model is prompted to identify findings in the image from a list of diseases.

Our task set also includes fine-grained 
tasks requiring detailed analysis of specific CXR regions.
These tasks involve grounding and 
identification of
findings, phrases, organs, and anatomical structures, as well as abnormality detection.
We convert bounding box labels and clinical text reports into instruction-following data 
to help the model link CXR images with spatial information for detailed analysis.

\subsubsection{Visual Question Answering.}

To train M4CXR, we convert all the data 
into an instruction-following format using predefined templates. 
However, training the LLM solely on 
templates presents a risk of overfitting: the model 
might 
lose its inherent conversational abilities and respond only in a fixed format. 
Therefore, we also capitalize on VQA datasets 
which are
already in a conversational format,
expecting
the model to freely respond to a diverse range of questions.

\section{Experiments}

\begin{table*}[ht]
\centering
\scalebox{0.9}{
\begin{tabular}{ccccccccccc}
\toprule
\multirow{2}{*}{Model} & \multicolumn{5}{c}{CheXbert} & Rad-Graph & \multicolumn{2}{c}{BLEU} & ROUGE & Test set \\ 
\cmidrule(r){2-6}
                       
                       & mF1-14 & mF1-5 & MF1-14 & MF1-5 & eF1-14 & F1 & -1 & -4 & -L & size \\ 
\midrule

LLM-CXR$^\dagger$  & 36.0 & -    & 21.1 & -    & -    & -    & 9.2  & 1.5  & 16.2 & 3,530  \\
RaDialog       & -    & -    & 39.4 & -    & -    & -    & 34.6 & 9.5  & 27.1 & -  \\
METransformer$^\dagger$ & -    & -    & -    & -    & 31.1 & -    & 38.6 & 12.4 & 29.1 & 3,269  \\
DCL$^\dagger$      & -    & -    & -    & -    & 37.3 & -    & -    & 10.9 & 28.4 & -  \\
PromptMRG      & -    & -    & 38.1    & -    & 47.6 & -    & 39.8 & 11.2 & 26.8 & 3,858 \\  
LM-RRG         & -    & -    & -    & -    & 48.4 & -    & -    & 12.2 & 29.6 & -  \\
Med-PaLM M 84B & 53.6 & 57.9 & 39.8 & \textbf{51.6} & -    & \textbf{26.7} & 32.3 & 11.5 & 27.5 & 4,834 \\
CheXagent$^*$  & 39.3 & 41.2 & 24.7 & 34.5 & -    & - & 16.9 & 4.7 & 21.5 & 2,461 \\
MAIRA-1    & 55.7 & 56.0 & 38.6 & 47.7 & -    & 24.3 & \textbf{39.2} & 14.2 & 28.9 &  2,461 \\
LLaVA-Rad  & 57.3 & 57.4 & 39.5 & 47.7 & -    & - & 38.1 & \textbf{15.4} & \textbf{30.6} & 2,461 \\

\midrule 
\multirow{2}{*}{M4CXR} & 58.1  & 61.6 & 38.8         & 49.5 & 50.2 & 21.7 & 33.3 & 10.2 & 28.4 & 3,858 \\
                              & \textbf{60.6}     & \textbf{61.8}   & \textbf{40.0} & 49.5 & \textbf{53.6}    & 21.8 & 33.9 & 10.3 & 28.5 & 2,461 \\ 
\bottomrule
\end{tabular}

}
\caption{
\textit{Single-image} MRG performance on the MIMIC-CXR test set.
The best performances are highlighted in bold. 
The numbers for CheXagent$^{*}$ were obtained from \citet{llava-rad}.
Models marked with $^\dagger$ report classification F1 scores 
using
a different labeler, CheXpert \cite{chexpert}.
}
\label{tab:result_singleview}
\end{table*}

\subsection{Training Datasets}
We collect and integrate various datasets according to the tasks described 
in the \textbf{Methods} section.

\subsubsection{Report Generation.}

MIMIC-CXR \cite{mimic-cxr} encompasses a diverse collection of CXR images along with detailed radiology reports. 
These reports are extensively annotated, facilitating advanced medical image analysis. 
For 
MRG tasks,
our focus is 
on generating the FINDINGS section, which offers an in-depth description of significant observations identified in the 
CXR images.

We employ CheXbert \cite{chexbert} to extract observation labels from the FINDINGS section.
CheXbert outputs 4-class (positive, negative, uncertain, blank) classification results for 14 predefined observation labels.
For binary classification, all non-positive classes are treated as negative.
Its label set of observations is provided as the candidate findings in the first question of CoT prompting.

For the three different MRG scenario tasks, image-report pairs are organized as follows.
For \textit{single-image}, every image is used as a data instance.
In \textit{multi-image}, 
images from different views that share the same study ID are gathered to compose study-level data. 
\textit{Multi-study} combines
two consecutive studies of a patient, 
with the studies arranged in chronological order.

\subsubsection{Image Understanding.}
BRAX \cite{brax} and CheXpert \cite{chexpert} datasets are used for disease classification.
We incorporate
datasets that contain bounding boxes along with class labels (VinDr-CXR \cite{vindr-cxr}, ChestX-ray14 \cite{nih}, ChestX-Det10 \cite{chestX-det10}, JSRT \cite{jsrt}, SIIM \cite{siim}, RSNA \cite{rsna}, COVID-19 Radiography \cite{covid-19}, COVID-QU-Ex \cite{qu-ex, qu-ex_2, covid-19}, QaTa-COV19 \cite{qata}) 
for various fine-grained image understanding tasks in addition to disease classification.
MS-CXR \cite{ms-cxr} is a dataset that includes 
image–sentence pairs of bounding boxes and corresponding phrases 
for images and reports from MIMIC-CXR.
ImaGenome \cite{imagenome}, which is also derived from MIMIC-CXR, annotates text in the report with bounding boxes aligned with 29 anatomical regions.

\subsubsection{Visual Question Answering.}
MIMIC-CXR-VQA \cite{mimic-cxr-vqa} and MIMIC-Diff-VQA \cite{mimic-diff-vqa} datasets are designed to address a wide range of questions based on MIMIC-CXR images.
These datasets encompass diverse and comprehensive question-answer pairs, enabling models to effectively handle various 
CXR-related queries.
RaDialog \cite{radialog} is a visual instruction-following dataset designed to facilitate tasks that require structured dialogues.
It guides models in understanding and responding to instructions based on CXR images.

\subsection{Test Datasets}

The test set used for evaluating 
all MRG tasks 
is from MIMIC-CXR.
The official test split 
includes 2,461 images in the frontal (PA, AP) view and 1,397 in other views, totaling 3,858 images.
Each study corresponding to the 2,461 frontal images is considered a separate data instance in \textit{multi-image} and \textit{multi-study}, with the prior study included for the latter if available.
As a result, the test set size for \textit{multi-image} and \textit{multi-study} is 2,461, while \textit{single-image} considers two test set sizes: 2,461 and 3,858.

For phrase grounding evaluation, the test set of MS-CXR is used.
Each query phrase in MS-CXR corresponds to a single bounding box.  
To ensure a rigorous evaluation of the task, we exclude all MS-CXR images from the training set of other datasets across all tasks.
Following the data split of \citet{medrpg}, the test set has 167 images.

We employ the test set of MIMIC-CXR-VQA for evaluating medical VQA.
Of the 13,793 total samples, we exclude 2,484 samples that lack answers, resulting in 11,309 samples for assessment. 
Additionally, we incorporate SLAKE \cite{slake}, which is not included in the training data. 
From its test set of 2,094 samples, we use only CXR images with close-ended questions in English, retaining 114 samples.

\subsection{Evaluation Metrics}
\subsubsection{Report Generation.}
The generated reports are assessed 
using
natural language generation (NLG) and clinical metrics.
We use the traditional BLEU \cite{bleu} and ROUGE-L \cite{rouge} for NLG metrics.

Regarding clinical accuracy, 
we calculate F1 scores from CheXbert classification results.
Macro-averaged F1 (MF1) and micro-averaged F1 (mF1) as well as example-based average F1 (eF1) scores are computed.
F1-14 indicates the F1 score calculated over all 14 CheXbert labels, while F1-5 is computed for only 5 of those labels (cardiomegaly, edema, consolidation, atelectasis, pleural effusion).

RadGraph-F1 \cite{radgraph-f1} is another 
clinical accuracy metric.
RadGraph \cite{radgraph} extracts a graph composed of clinical entities and relations from a radiology report. 
RadGraph-F1 is the average of F1 scores calculated for both the entities and relations of the graph.

\subsubsection{Medical Phrase Grounding.}
To evaluate the performance of phrase grounding, we calculate the intersection over union (IoU) between the ground-truth and predicted bounding boxes.
We use the mean IoU (mIoU) averaged over all data as the evaluation metric.
Additionally, accuracy is determined by considering predicted boxes with an IoU of 0.5 or higher as correct predictions.

\subsubsection{Visual Question Answering.}
Accuracy, recall, and BLEU-1 are used to assess the VQA performance.
Exact matches between predictions and ground-truth are counted to calculate the accuracy.
To consider partial matches for open-ended responses, recall is calculated by measuring the proportion of ground-truth words present in the generated sequences.
For MIMIC-CXR-VQA, which includes open-ended questions, BLEU-1 scores are also calculated for each test sample and then averaged.

\subsection{Model Training}
We use a randomly initialized C-Abstractor as the projector
to efficiently handle multiple radiology images and reduce 
image tokens.
The LLM and vision encoder are pre-trained models:
Mistral-7B-Instruct-v0.2 \cite{mistral} and Rad-DINO \cite{raddino}, respectively.
The training 
involves two stages.
First, we pre-train the projector while keeping the LLM and vision encoder frozen, using only CXR images and reports without instruction prompts. 
Subsequently, the vision encoder, projector, and LLM are trained together for visual instruction tuning. 
We apply LoRA \cite{lora} for LLM training to reduce computational cost.
Each instruction-following data input includes images and instruction texts, and the model is trained to predict the corresponding responses using cross-entropy loss.

The sampling ratio of datasets for each task significantly impacts 
M4CXR's performance.
The empirical ratio for multi-tasking capabilities was determined
to be approximately 54\%, 35\%, and 11\% for MRG, image understanding, and VQA, respectively.
The detailed training hyperparameters (Table \ref{tab:hyperparams}),
specific sampling ratios (Table \ref{tab:ratio}),
and the exploration of sampling ratios
are provided in the Appendix.

\section{Results}

\subsection{Medical Report Generation}
\subsubsection{Single-Image.}

To evaluate \textit{single-image} MRG performance, we compared our model with state-of-the-art MRG models, including 
LLM-CXR, RaDialog, 
METransformer \cite{metransformer}, DCL \cite{dcl}, PromptMRG, LM-RRG \cite{lmrrg}, 
CheXagent, MAIRA-1 \cite{maira1}, Med-PaLM M, and LLaVA-Rad \cite{llava-rad}.
The results 
are summarized in Table \ref{tab:result_singleview}.
For direct comparison with other models, 
we evaluated M4CXR using two distinct test sets: 
one with 2,461 frontal images
and another with
3,858 images of all views.
Including lateral views 
decreased
clinical accuracy, 
suggesting the complexity of recognizing observations from lateral images.

Among the models 
evaluated on only frontal images, 
M4CXR
attained the highest 
CheXbert clinical accuracy, 
with mF1-14 and MF1-14 scores of 60.6 and 40.0, respectively. 
In the evaluation that included all views, 
our model outperformed PromptMRG with an eF1-14 of 50.2.
Although Med-PaLM M used a different all-view test set, which limits direct comparison, our model showed a significant advantage in mF1, with similar or slightly lower MF1s.

However, M4CXR did not achieve the best scores in both RadGraph-F1 and NLG metrics. 
To improve clinical accuracy, CoT prompting provides candidate findings in the first question before generating reports.
This approach may have led the model to use terms from the findings list
rather than words that exactly match those in the ground-truth report, 
which could be a contributing factor to the observed lower NLG scores.
RadGraph-F1, calculated by extracting entities and relations from the report, 
was likely reduced due to this mismatch.

\subsubsection{Various Scenarios.}
Table \ref{tab:scenarios} shows the evaluation results across various input scenarios. 
\textit{Multi-image} improved clinical accuracy compared to \textit{single-image}, 
with CheXbert mF1-14 and MF1-14 scores of 61.1 and 41.0, respectively.
Providing the prior study as additional input
improved MF1, though mF1 slightly decreased.
These results demonstrate that our model can effectively utilize 
available inputs 
to generate medical reports in various scenarios.

Although test conditions differ, we compared our model's performance with
two baseline models.
 \cite{medpalm-m} reported the zero-shot generalization results of Med-PaLM M
in a two-view setting.
Since it was trained only on \textit{single-image}, its effectiveness diminished, 
unlike M4CXR leveraging
additional images for enhanced outcomes.
MAIRA-2 accepts a prior study and multiple images as input, akin to our \textit{multi-study} scenario.
Its performance matches ours closely: MAIRA-2 achieved a higher MF1-14 of 42.7, whereas M4CXR surpassed it with an mF1-14 of 60.7.

\begin{table}[ht]
\centering
\scalebox{0.9}{
\begin{NiceTabular}{c|ccc}
\toprule
\multirow{2}{*}{Model} & \multirow{2}{*}{MRG Scenario} & \multicolumn{2}{c}{CheXbert} \\ \cline{3-4} 

                       &                                      & mF1-14     & MF1-14    \\ 
                       \midrule
Med-PaLM M 84B             & \textit{Multi-image}            & 50.5         & 37.8     \\   
MAIRA-2 7B                   & \textit{Multi-study}           & 58.5         & \textbf{42.7}     \\   
\midrule 
\multirow{3}{*}{M4CXR}& \textit{Single-image}           & 60.6         & 40.0     \\ 
                             & \textit{Multi-image}            & \textbf{61.1}         & 41.0     \\ 

                             & \textit{Multi-study}           & 60.7         & 42.0     \\ 
                             \bottomrule
\end{NiceTabular}
}
\caption{
Evaluation results of MRG performance in various scenarios.
}
\label{tab:scenarios}
\end{table}

\subsection{Medical Phrase Grounding}

Table \ref{tab:result_ground}
compares the medical phrase grounding results of TransVG \cite{transvg}, MedRPG, MAIRA-2, and M4CXR, on the MS-CXR test set. 
While MedRPG achieved the highest accuracy and mIoU, 
MedRPG and TransVG are 
specialized for phrase grounding and cannot perform other tasks.
Given this limitation, M4CXR, 
a multi-tasking model,
shows 
competitive performance
with an accuracy of 68.3 and mIoU of 57.9.
MAIRA-2, 
utilizing a private dataset for grounded report generation, 
reported a similar mIoU to ours.

\begin{table}[ht]
\centering
\scalebox{0.9}{
\begin{NiceTabular}{c|cc}
\toprule
Model                    & Accuracy            & mIOU     \\ 
\midrule
TransVG                  & 65.9          & 58.9    \\ 
MedRPG                   & \textbf{69.9}          & \textbf{59.4}    \\ 
MAIRA-2 7B        & -              & 57.8    \\ 
\midrule 
M4CXR             & 68.3          & 57.9    \\ 
\bottomrule
\end{NiceTabular}
}
\caption{
Referring expression grounding results. 
Note that MAIRA-2 7B was evaluated on a different test split.
}
\label{tab:result_ground}
\end{table}

\subsection{Visual Question Answering}

Table \ref{tab:result_vqa} 
presents the results of medical VQA evaluation. 
We conducted our own evaluations on 
open-source
medical MLLMs: LLaVA version of RaDialog \cite{radialog_1.1.0}, RadFM \cite{radfm}, and CheXagent.
M4CXR outperformed the other models, except for the recall on MIMIC-CXR-VQA.
CheXagent included the dataset in its training, likely explaining its high recall of 72.8.
The lower accuracy on MIMIC-CXR-VQA, compared to SLAKE, is due to its greater variety of questions.
Moreover,
the higher BLEU-1 score compared to accuracy suggests the presence of partially correct answers.

\begin{table}[ht]
\centering
\scalebox{0.9}{
\begin{NiceTabular}{c|ccc|cc}
\toprule
\multirow{2}{*}{Model}  & \multicolumn{3}{c|}{MIMIC-CXR-VQA} & \multicolumn{2}{c}{SLAKE} \\
\cmidrule(lr){2-4} \cmidrule(l){5-6} 
    & Acc & Recall  & BLEU-1    & Acc & Recall \\ 
\midrule
RaDialog & 0.0 & 43.0 & 5.0 & 0.0 & 45.6\\
RadFM & 11.2& 36.7& 13.8& 68.4 & 69.7\\
CheXagent & 59.0 & \textbf{72.8} & 62.5 & 71.1 & 73.2 \\ 
\midrule 
M4CXR & \textbf{62.3} & 70.3 & \textbf{66.4} & \textbf{85.1} & \textbf{86.0}    \\ 
\bottomrule
\end{NiceTabular}
}
\caption{
Medical VQA performance. 
}
\label{tab:result_vqa}
\end{table}

\begin{table*}[ht!]
\renewcommand{\arraystretch}{1.1}
\centering
\scalebox{0.9}{
\begin{NiceTabular}{c|ccc|c|cccccc}
\toprule
                       & \multicolumn{3}{c|}{Training}           & Test   & \multicolumn{2}{c}{CheXbert}    & \multicolumn{2}{c}{NLG} & Grounding & VQA
                       \\ \cline{2-11}                                      
                       & MRG & ImgUnd & VQA & CoT & mF1-14 & MF1-14 & BLEU-4 & ROUGE-L & mIoU & BLEU-1 \\
\midrule
\multirow{2}{*}{M4CXR}    & \multirow{2}{*}{\ding{51}}         & \multirow{2}{*}{\ding{51}} & \multirow{2}{*}{\ding{51}} & \ding{51} & \textbf{60.6}& \textbf{40.0} & \textbf{10.3}  & \textbf{28.5} & \multirow{2}{*}{\textbf{57.9}}  & \multirow{2}{*}{66.4}\\ 
                             &                                    &                            &                            & \ding{55} & 49.2 & 31.2 & 9.3   & 27.3 &    &   \\           \midrule 
Exp1                         & \ding{51}$^{s}$                    & \ding{51}                  & \ding{51}                  & \ding{55} &  50.0 & 33.1 & 9.3 & 27.7 & 56.1  & 66.1
\\ \midrule
\multirow{2}{*}{Exp2}        & \multirow{2}{*}{\ding{51}$^{s,c}$} & \multirow{2}{*}{\ding{51}} & \multirow{2}{*}{\ding{51}} & \ding{51} & 58.5 & 37.5 &  0.6 & 10.9 & \multirow{2}{*}{56.3} & \multirow{2}{*}{66.6} \\ 
                             &                                    &                            &                            & \ding{55} & 49.2 & 31.6 & 9.2  & 27.2 &  &  \\  \midrule  
Exp3                         & \ding{51}                          &                            & \ding{51}                  & \ding{51} & \textbf{60.6} & 38.2 & 10.1 & 28.2 & - & \textbf{67.9}  \\  \midrule 
Exp4                         & \ding{51}                          & \ding{51}                  &                            & \ding{51} & 59.3 & 38.3 & 10.1 & 28.0 & \textbf{57.9} & 36.1  \\  
\bottomrule
\end{NiceTabular}
}
\caption{
Results of ablation studies.
``ImgUnd'' stands for ``Image Understanding''.
\ding{51}$^{s}$ and \ding{51}$^{s,c}$ indicate single-turn MRG without CoT prompting and 
the training of single-turn MRG and disease classification as separate tasks, respectively.
}
\label{tab:ablation}
\end{table*}

\begin{figure*}[ht]
\centering
\includegraphics[width=0.95\textwidth]{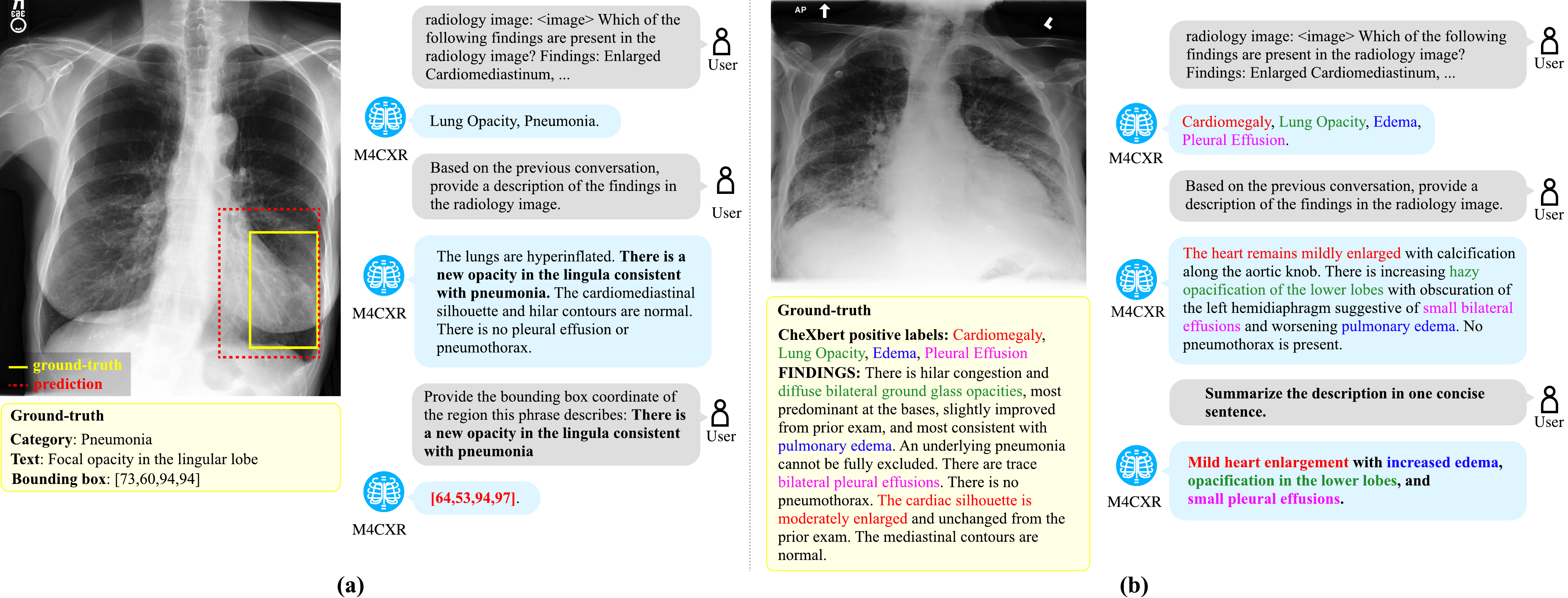}
\caption{
Examples of M4CXR's performance in (a) visual grounding and (b) VQA.
The images are selected from the test splits of MS-CXR and MIMIC-CXR, respectively.
}
\label{fig:analysis}
\end{figure*}

\subsection{Ablation Study}

We examined the effects of multi-turn CoT prompting and 
the combination of task types 
through comparative experiments (Table \ref{tab:ablation}).
Generating reports
without CoT prompting 
in M4CXR
led to a significant drop in clinical accuracy.
Exp1, trained on single-turn MRG without the first question of CoT prompts,
yielded similar results.
Exp2 involved training on 
single-turn MRG and disease classification
as separate tasks. 
Testing Exp2 with CoT prompts
improved clinical accuracy, 
but
NLG metrics fell significantly,
with the generated text being
merely a list of identified observations.
The experiments reveal 
the effectiveness of multi-turn CoT prompting in 
MRG's training and testing phases.

Exp3 and Exp4 
excluded
image understanding and VQA tasks, respectively. 
In Exp3, while MRG performance was on par and VQA performance improved due to the increased proportion of VQA data, the absence of image understanding data precluded visual grounding.
In Exp4, the BLEU-1 score for VQA dropped significantly from 66.4 to 36.1, 
suggesting the necessity of VQA datasets 
in understanding and answering free-form questions.
Moreover, the slight decrease in Exp4’s clinical accuracy compared to M4CXR implies 
the contribution of VQA datasets to MRG performance improvement.
Detailed examples are in the Appendix.

\subsection{Qualitative Analysis}

Figure \ref{fig:analysis} shows two examples of dialogs between a user and M4CXR, each for visual grounding and VQA.
Additional examples can be found in the Appendix.
In Figure \ref{fig:analysis} (a), 
after the multi-turn CoT prompting to generate a report, the user asks a third question to locate the region of a sentence in the generated report.
The model provides the coordinates of a bounding box, depicted as a red dashed box.
The prediction covers all of the ground-truth, 
represented by 
the yellow solid box, 
possessing sufficient explanatory power to indicate what the corresponding sentence represents.

Figure \ref{fig:analysis} (b) illustrates M4CXR's ability to write a clear and concise summary from the generated report.
On the left, the ground-truth report is provided for comparison.
For 
enhanced 
visual clarity, 
findings and corresponding sentences are color-matched.
M4CXR accurately identifies all the observations in the first question, 
and subsequently generates a comprehensive report without omitting any detail.
In response to the third question which asks for a one-sentence summary, the model 
offers a concise statement 
that includes every noted finding.
A notable flaw is 
that 
expressions such as ``increasing'' and ``worsening,'' 
which would be meaningful only
in a \textit{multi-study} scenario, 
were generated despite the \textit{single-image} context.
While it is understandable that this hallucination is due 
to 
the ground-truth reports containing such expressions 
during training, 
it is evident that there is room for improvement.

\section{Conclusion}

In this study, we introduced M4CXR, a multi-modal LLM aimed at enhancing CXR interpretation by leveraging the versatile advantages of LLMs.
Trained on a CXR visual instruction-following dataset constructed 
by appropriately combining various datasets, 
M4CXR is capable of performing multiple tasks.
Exploiting its reasoning capability, we proposed a novel chain-of-thought prompting strategy that significantly improved the clinical accuracy
of report generation.
By identifying observations from X-ray images and generating descriptions through multi-turn conversations,
M4CXR achieved notable improvements in
CheXbert F1 scores
compared to existing models.
The model's ability to handle multiple images and incorporate prior studies demonstrated its adaptability in diverse clinical scenarios.

Beyond report generation, M4CXR also proved to be highly effective in visual grounding and VQA.
M4CXR competed well with models specialized for visual grounding,
while also exhibiting outstanding performance in VQA.
Qualitative analysis highlights M4CXR's flexibility in answering free-form questions.
With future work to address limitations such as 
low NLG metrics and the presence of hallucinations,
we anticipate 
that further advancements
in M4CXR will contribute to
the development of a 
highly capable radiology assistant.

\bibliography{aaai25}

\clearpage

\section{Appendix}

\subsection{Implementation Details}
The detailed hyperparameters for training are summarized in Table \ref{tab:hyperparams}. 
C-Abstractor 
is configured with a depth of 3, an MLP depth of 2, and a hidden size of 1024, to be employed as the projector.
The number of visual tokens is set to 361, corresponding to a 19 $\times$ 19 grid.
Accordingly, the output of the vision encoder, which is a sequence of 1369 (37 $\times$ 37) tokens processed by Rad-DINO,
is compressed to a length of 361.
This reduction in the number of tokens, 
by approximately one-fourth,
enables the LLM to handle a greater number of images more efficiently.
In the training of the LLM, we configure the LoRA hyperparameters as follows: rank is set to 8, alpha to 32, and dropout to 0.05.
Experiments are conducted on two H100 GPUs,
using bfloat16 with automatic mixed precision 
and flash-attention v2 \cite{flash-attention-2}.
For evaluation,
sentences are generated with greedy search decoding.
One exception is the qualitative evaluation of VQA: to generate more natural responses, multinomial sampling is applied.

\begin{table}[ht]
\centering
\scalebox{0.9}{
\begin{NiceTabular}{c|c|c}
\toprule
Configuration            & Pre-training & Instruction tuning                    \\ 
\midrule
\midrule
\multirow{2}{*}{Training module}  & \multirow{2}{*}{Projector} & Vision Encoder, \\ 
                                  &           & Projector, LLM \\ 
\midrule
Training steps      & 2k           &                40k                        \\ \midrule
Max steps           & 10k           &                100k                        \\ \midrule
Warmup steps        & \multicolumn{2}{c}{500}                                \\ \midrule
Global batch size   & 256          & 64                                      \\ 
\midrule
Optimizer           & \multicolumn{2}{c}{AdamW}                              \\ 
\midrule
Optimizer      & \multicolumn{2}{c}{\multirow{2}{*}{$\beta_{1}=0.9$, $\beta_{2}=0.999$, $\epsilon=1e^{-8}$}}        \\ 
hyperparameter & \multicolumn{2}{c}{}                                                                               \\
\midrule
Learning rate            & 3e-4         &               1e-4                         \\ 
\midrule
Learning rate schedule   & \multicolumn{2}{c}{Cosine decay}                             \\ 
\midrule
Weight decay             & 1e-2         & 1e-4                                    \\ 
\midrule
Gradient clipping                & \multicolumn{2}{c}{1.0}                              \\ 
\bottomrule

\end{NiceTabular}
}
\caption{Training hyperparameters.}
\label{tab:hyperparams}
\end{table}

\subsection{Details on Chest X-ray Interpretation Tasks}
Detailed descriptions for all the CXR interpretation tasks considered in this work are listed in Table \ref{tab:dataset-description}.
Table \ref{tab:dataset-tem} provides the conversation templates for each task. 
Although not shown in the table, a system prompt, which specifies that the AI medical assistant should give helpful and detailed answers, is inserted before the first question in each template.
Table \ref{tab:ratio} lists the datasets used for each task, along with the number of training instances and the sampling ratios applied during training.

\subsection{Dataset Preprocessing}
We follow the official split for every dataset. 
The additional preprocessing steps, taken for each dataset as required, are described in detail.

\subsubsection{MIMIC-CXR.}
We extract the FINDINGS section using the official preprocessing code\footnote{https://github.com/MIT-LCP/mimic-cxr/tree/master/txt}. 
Then, we 
use the text preprocessing code from \citet{r2gen} 
with slight modification 
to remove special characters, numbering, and extra spaces. 
We exclude reports that lack a FINDINGS section or have a length of 
less than 5 characters.
For \textit{multi-image}, we consider only studies with 
at most 5
images. 
For \textit{multi-study}, we include only cases where the combined number of images in the prior and follow-up studies is 
at most 10.
Out of 2,461 follow-up studies in the test set, 
only 2,333 studies have a prior study and the remaining 128 studies have no preceding study.

\subsubsection{VinDR-CXR.}
Due to the presence of many similar bounding box labels, overlapping boxes in an image are merged into the smallest possible bounding box if they overlap by more than 50 percent.
\subsubsection{JSRT.}
Circular labels are converted to the smallest possible bounding box labels.
\subsubsection{Mask Labeled Datasets.}
The mask labels 
are converted to bounding boxes that minimally enclose each mask.
This conversion 
is 
applied for 
ChestX-Det10, SIIM, COVID-19 Radiography, COVID-QU-EX, and QaTa-COV19.
\subsubsection{COVID-19 Radiography.}
Images with more than three mask regions are excluded.
\subsubsection{QaTa-COV19.}
Data that overlaps with COVID-QU-EX are removed.
\subsubsection{RSNA.}
Only `lung opacity' and `normal' labels are used. 
According to \citet{cxrclip}, `lung opacity' is referred to as `pneumonia'.
\subsubsection{ImaGenome.}
We utilize the preprocessing code from \citet{rgrg}. 
As a result, we obtain
report text, bounding boxes, and anatomical names for 29 anatomical regions.
\subsubsection{RaDialog.}
We use only the images from MIMIC-CXR.
Among the tasks introduced in \citet{radialog}, we exclude
RG (report generation).
We utilize version 1.1.0 \cite{radialog_1.1.0} of the dataset.

\subsection{Exploration of Sampling Ratios}
 
In multi-task learning, the data sampling strategy plays a crucial role, with 
numerous 
such strategies being applicable.
We experiment with various sampling strategies, as shown in Table \ref{tab:data_sampling_comparison}.
The tasks are grouped into three task types, and a dataset used within a single task is referred to as a task-dataset.
The per-task-dataset (D1) strategy samples uniformly across all task-datasets. 
The per-size (D2) strategy samples proportionally based on the size of each task-dataset. 
After setting equal proportions for each task type (per-task-type), within each task-type,
D3 samples per-task-dataset while D4 samples per-size.

In D1, which follows the per-task-dataset strategy, MRG accounts for only 3 out of 50 total task-datasets.
This very low sampling ratio for MRG explains its low MRG performance.
In contrast, in the per-task-type settings of D3 and D4, one-third of the overall tasks are allocated to MRG, leading to improved MRG performance compared to D1. 
For grounding performance, D2 and D4 showed inferior results, likely because the 
training split of MS-CXR, which is the dataset used for evaluation,
was barely sampled when using the per-size setting. 
As for VQA, since all datasets are large in size, 
per-size sampling (D2) exhibited a high BLEU-1 score. 
Similar to MRG, VQA also constitutes a small proportion of the total task-dataset (4 out of 50), resulting in lower performance for D1, while 
D3 and D4 achieved higher scores
as per-task-type allocates one-third to VQA.

We aimed to 
find the appropriate
sampling ratios where M4CXR remains clinically accurate, supports visual grounding, and retains the conversational abilities of the LLM.
To improve MRG performance, we reduced the ratio allocated to VQA and increased the proportion dedicated to MRG.
For 
image understanding, 
we began with the per-size ratio and made further adjustments to enhance grounding performance.
Additionally, 
to improve the instruction-following capabilities of the model,
we increased the ratio of
RaDialog compared
to other VQA datasets. 
While this adjustment led to a decrease in quantitative VQA performance metrics, it 
contributed in preserving
the LLM's ability to understand and respond to instructions in conversation.

\subsection{Qualitative Examples}

\subsubsection{Comparison of MRG Scenarios.}

\begin{figure*}[ht]
\centering
\includegraphics[width=0.9\textwidth]{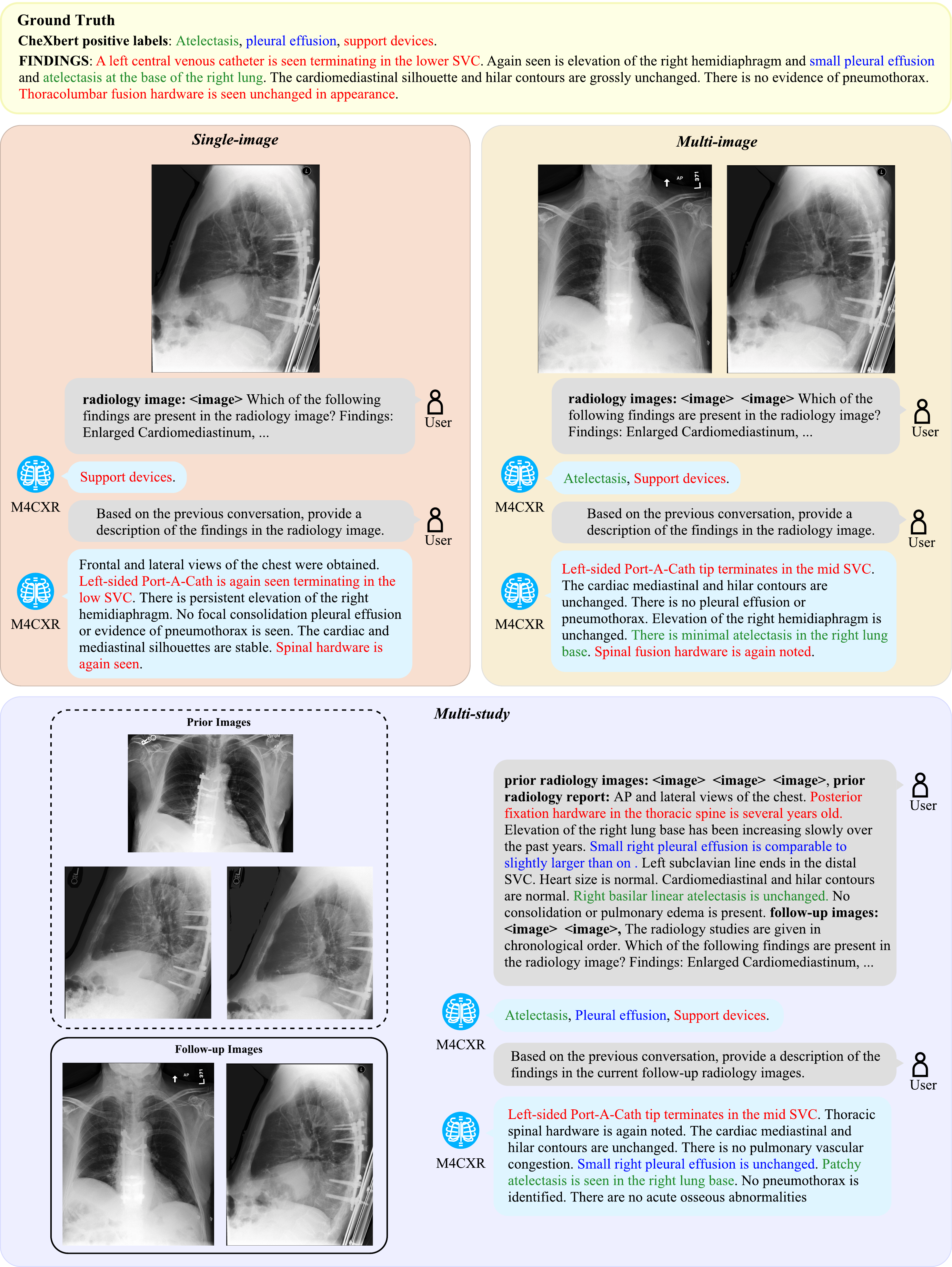}
\caption{
Examples of medical report generation across various scenarios. 
For the same study, the top left shows the result for \textit{single-image}, the top right for \textit{multi-image}, and the bottom for \textit{multi-study} report generation. 
The results are selected from the MIMIC-CXR test set.
}
\label{fig:senario_appendix}
\end{figure*}

Figure \ref{fig:senario_appendix} presents the results of 
report generation
for the same study across various scenarios, highlighting how clinical accuracy improves as more information becomes available. 
The ground-truth targets, including atelectasis, pleural effusion, and support devices, are each marked in different colors.
In \textit{single-image}, one lateral image is used; 
in \textit{multi-image}, 
both AP and lateral images
are used; 
and in \textit{multi-study}, three prior images and the corresponding report are included.

In \textit{single-image}, only the support devices are correctly identified, suggesting that a single lateral image may not provide sufficient information for accurate CXR interpretation. 
In \textit{multi-image}, the addition of a frontal image enables the correct identification of atelectasis. 
In \textit{multi-study}, all three targets are successfully identified, 
possibly due to the prior report already containing 
these three findings.
This demonstrates that M4CXR is capable of understanding and utilizing the information provided in various MRG scenarios.

\subsubsection{Visual Grounding.}

\begin{figure*}[ht]
\centering
\includegraphics[width=0.95\textwidth]{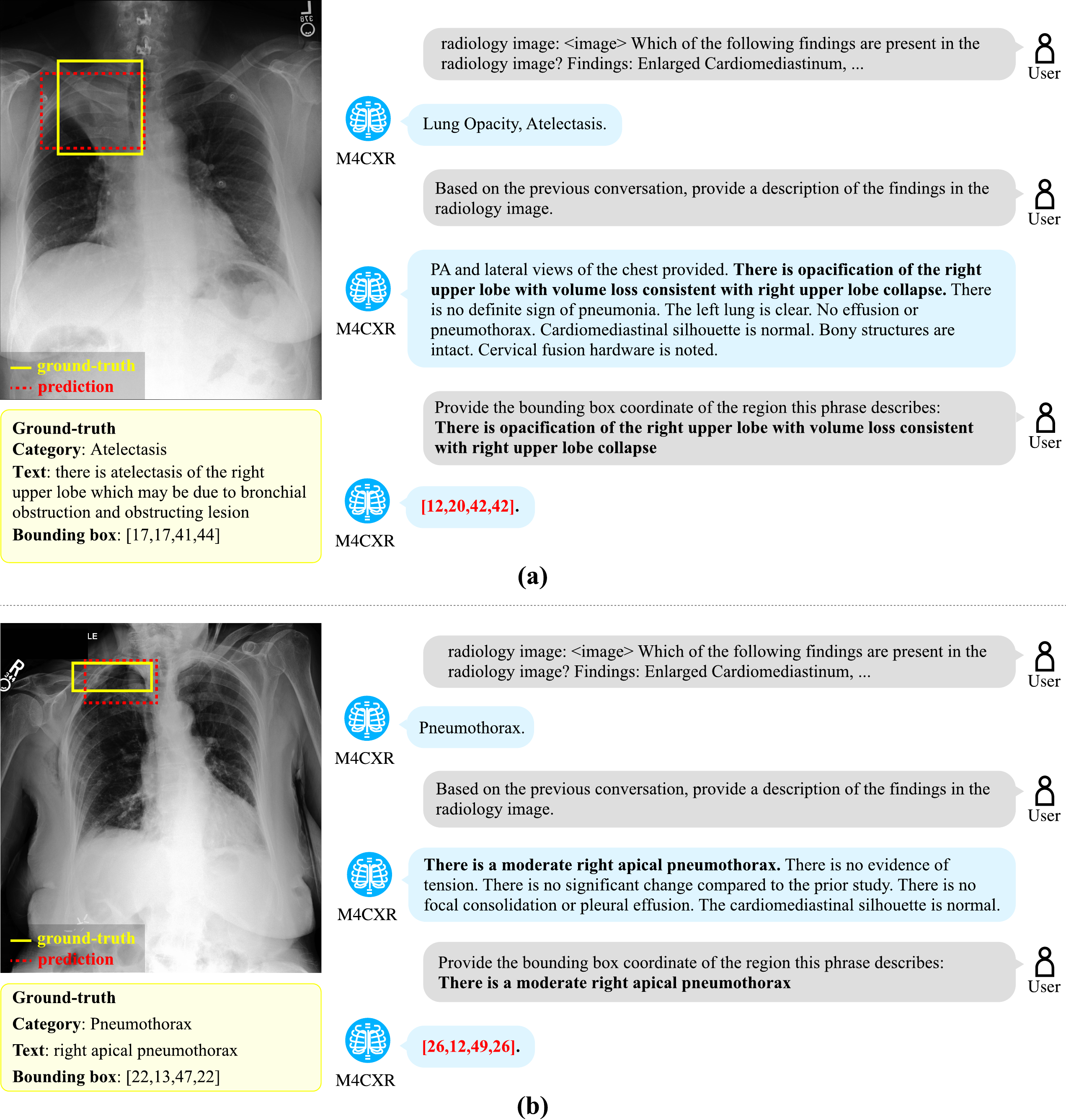}
\caption{Examples of visual grounding. 
The ground-truth bounding box is represented by 
a yellow solid box, 
while the prediction is shown with 
a red dashed box. 
The images are selected from the MS-CXR test set.
}
\label{fig:grounding_appendix}
\end{figure*}

Figure \ref{fig:grounding_appendix} presents additional examples of visual grounding. 
In both (a) and (b), 
the user asks a third question after MRG
to identify the location referenced by a specific phrase, 
and M4CXR responds with the coordinates of a bounding box.

In Figure \ref{fig:grounding_appendix} (a), the model predicts  
the presence of lung opacity and atelectasis. 
The report contains the phrase ``volume loss consistent with right upper lobe collapse'', which suggests atelectasis. 
When asked to ground 
the sentence containing this phrase,
the model 
identifies an area that largely overlaps with
the ground-truth bounding box.
In Figure \ref{fig:grounding_appendix} (b), the generated report explicitly mentions the presence of pneumothorax, and the model correctly grounds the upper part of the right lung.
These examples demonstrate that M4CXR can be utilized both for generating reports and for determining the locations referenced by the generated text.

\subsubsection{Impact of VQA Datasets.}

\begin{figure*}[ht]
\centering
\includegraphics[width=0.95\textwidth]{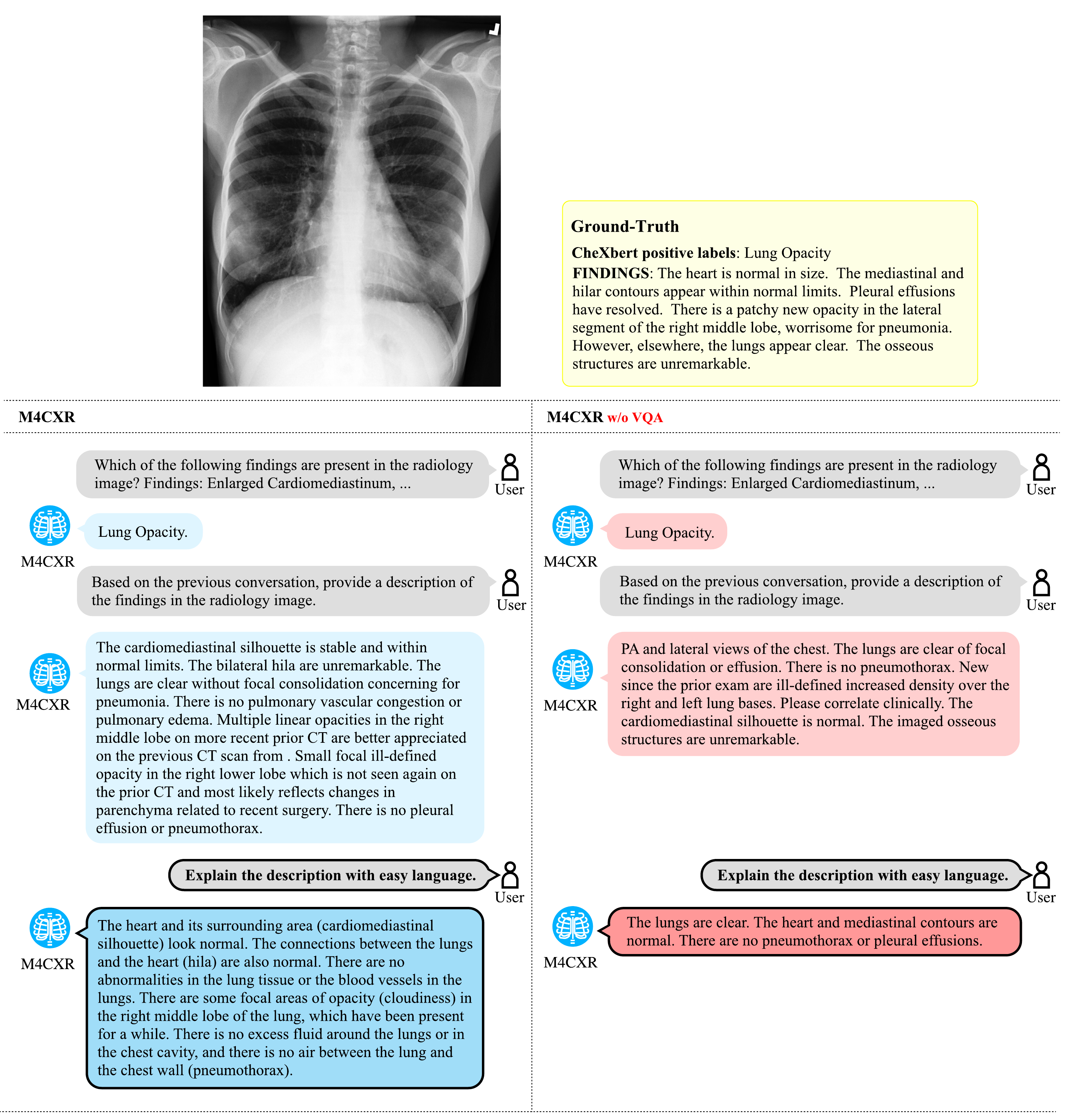}
\caption{
Examples of medical report explanations using easy language. 
The left shows the results from M4CXR, while the right shows the results when trained without VQA datasets.
}
\label{fig:vqa_appendix_1}
\end{figure*}

\begin{figure*}[ht]
\centering
\includegraphics[width=0.95\textwidth]{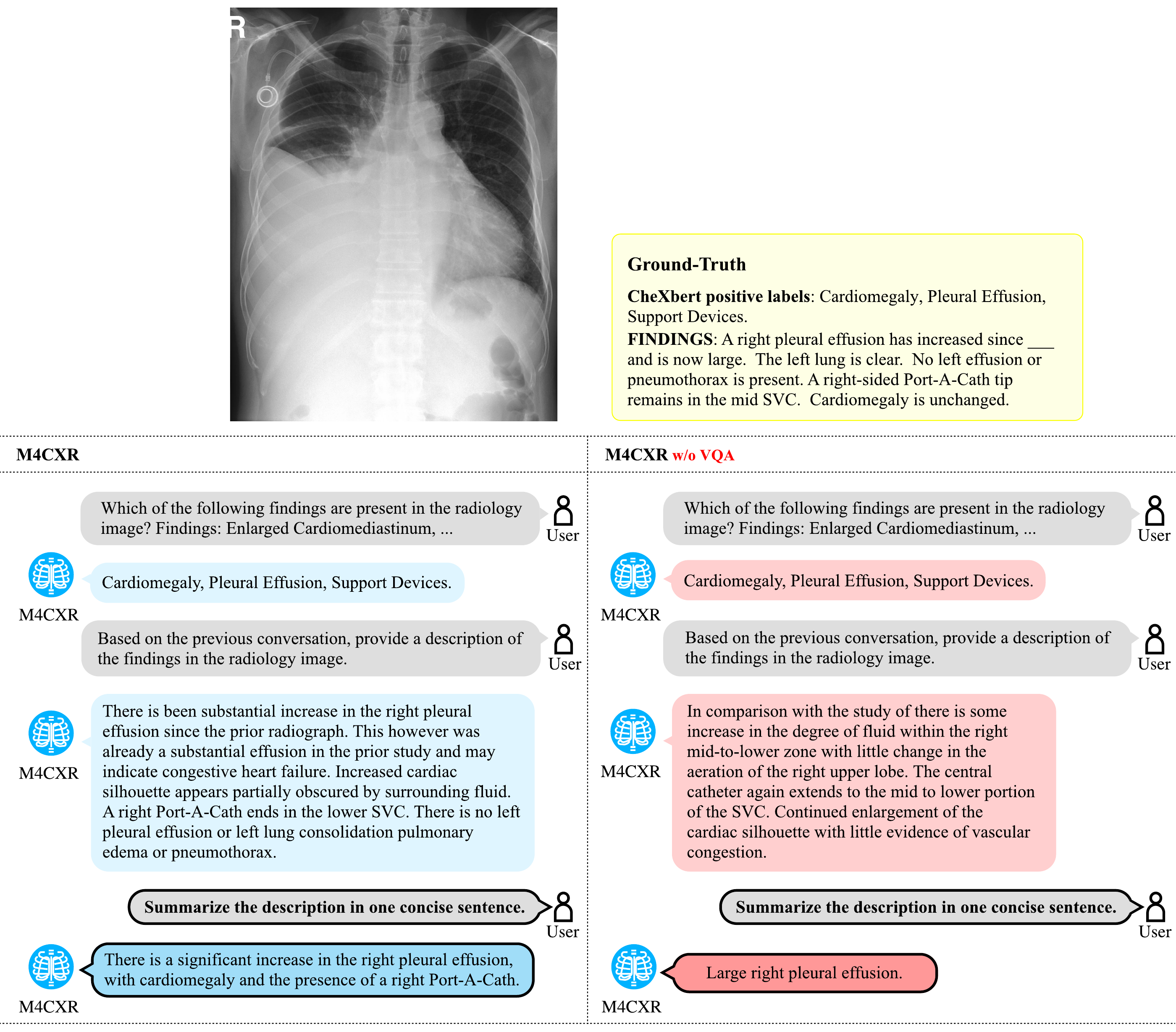}
\caption{
Examples of medical report summarization. 
The left shows the results from M4CXR, while the right shows the results when trained without VQA datasets.
}
\label{fig:vqa_appendix_2}
\end{figure*}

Figures \ref{fig:vqa_appendix_1}, \ref{fig:vqa_appendix_2}, and \ref{fig:vqa_appendix_3} collectively illustrate the differences in VQA performance 
between M4CXR and the
Exp4 model from \textbf{Ablation Study}, emphasizing the impact of incorporating 
VQA datasets.
The generated responses were compared using images selected from the MIMIC-CXR test set.

In Figure \ref{fig:vqa_appendix_1}, when 
requested to rephrase the generated reports
in easy language, M4CXR effectively translates medical jargon into general language. 
In contrast, Exp4 model merely summarizes the report content, continuing to use medical terminology without making it easier to understand.

Figure \ref{fig:vqa_appendix_2} presents the comparison results for 
the task of report summarization. 
M4CXR successfully condenses the report into a concise, single sentence as requested,
notably including all important observations 
such as cardiomegaly, pleural effusion, and support devices. 
In contrast, the model trained without VQA datasets oversimplifies the content, mentioning only pleural effusion and omitting other critical details.

Figure \ref{fig:vqa_appendix_3} shows the responses when recommendations for diagnosis and treatment are requested. 
M4CXR suggests continued observation and follow-up imaging studies based on the findings, and also recommends treatments such as pleural drainage or chest tube placement if the condition worsens.
In contrast, 
Exp4 model
provides a simpler response, suggesting evaluation with CT and the need for monitoring.

These comparisons between the two models demonstrate
that VQA datasets play a significant role in maintaining the LLM’s ability to 
follow instructions and provide appropriate responses.

\subsubsection{Hallucination.}
\begin{figure*}[ht]
\centering
\includegraphics[width=0.95\textwidth]{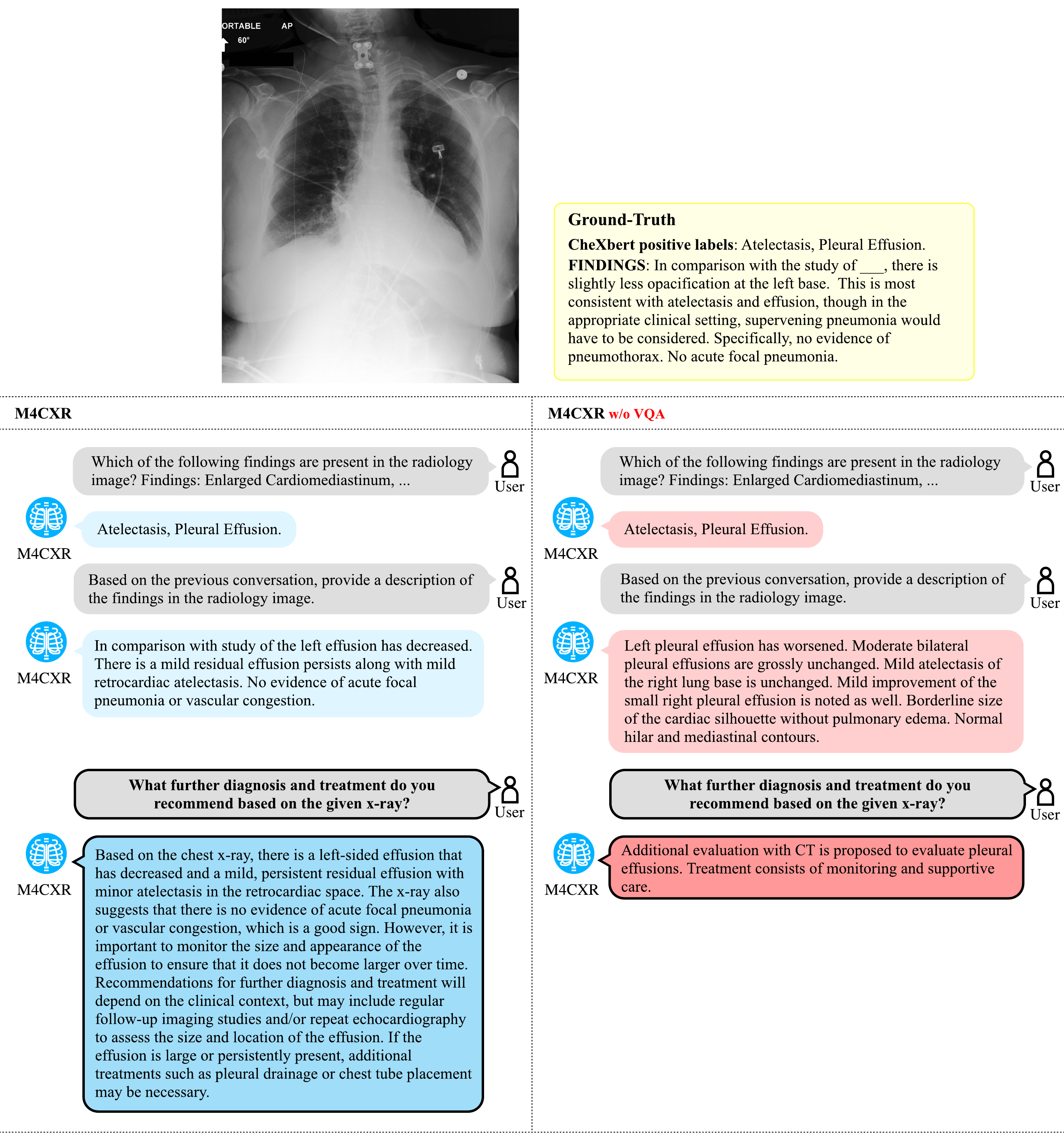}
\caption{
Examples of medical treatment recommendation. 
The left shows the results from M4CXR, while the right shows the results when trained without VQA datasets.
}
\label{fig:vqa_appendix_3}
\end{figure*}

In Figure \ref{fig:vqa_appendix_3}, the medical report generated by M4CXR contains comparative contents such as ``In comparison with study'' and ``decreased'', 
even though it was generated in a \textit{single-image} scenario. 
Similar issues can be observed in other examples, where the model often refers to comparisons or mentions images that 
are not provided.
During the training of \textit{single-image} MRG, the use of certain ground-truth reports, those written specifically in a \textit{multi-study} context, is likely to have induced these hallucinations.
To address this issue, ground-truth reports tailored to each MRG scenario are needed, 
which
could be explored in future work.

\begin{table*}[ht]
\renewcommand{\arraystretch}{1.1}
\centering
\scalebox{0.9}{
\begin{NiceTabular}{c|c|cccccc}
\toprule
 \multirow{2}{*}{}   & \multirow{2}{*}{Sampling Strategy}   & \multicolumn{2}{c}{CheXbert}    & \multicolumn{2}{c}{NLG} & Grounding & VQA \\
 \cline{3-8}  
                       &  & mF1-14 & MF1-14 & BLEU-4 & ROUGE-L & mIoU & BLEU-1 \\
\midrule
M4CXR                   & empirical ratio            &  \textbf{60.6} & \textbf{40.0} & \textbf{10.3} & \textbf{28.5} & 57.9 & 66.4 \\  \midrule 
D1               & per-task-dataset    &  52.7 & 32.1 &  9.1 & 26.8 & \textbf{58.5} & 72.8 \\
D2               & per-size &  58.5 & 36.1 & 9.0  & 27.2 & 47.9  & 77.4 \\  
D3               & per-task-type \& per-task-dataset  &  59.0 & 37.1 & 10.0  & 28.2 & 57.9  & \textbf{78.3} \\ 
D4               & per-task-type \& per-size   &  59.1 & 37.7 & 9.1  & 27.3 & 46.4  & 78.0 \\
\bottomrule
\end{NiceTabular}
}
\caption{
Performance comparison between different data sampling strategies.
}
\label{tab:data_sampling_comparison}
\end{table*}

\begin{table*}[ht]
\centering
\renewcommand{\arraystretch}{1.2}
\scalebox{1.0}{
\begin{NiceTabular}{p{0.12\linewidth}|p{0.2\linewidth}|p{0.6\linewidth}}
\toprule
Task Type & Task & Description\\
\midrule
\midrule
\multirow{5}{=}{Medical Report Generation}  &  \textit{Single-image} & Generates the FINDINGS section of a report from a single CXR image. \\ 
\cline{2-3}
&  \textit{Multi-image}   & Generates the FINDINGS section of a report from one or more images, including images from different views within the same study.  \\ 
\cline{2-3}
&   \textit{Multi-study} & Generates the FINDINGS section of a report from one or more studies, combining current and previous studies from a patient.  \\ 
\midrule
\multirow{17}{=}{Image Understanding} &  Disease Classification & Identifies diseases within a single image based on the labels covered by each dataset. \\
\cline{2-3}
&   Finding Grounding & Provides the bounding box coordinates for a given finding, if detected.\\
\cline{2-3}
&  Grounded Finding & Identifies the finding corresponding to a given bounding box. \\
\cline{2-3}
&  Abnormality Detection & Identifies and localizes abnormal regions by providing the corresponding bounding box coordinates.\\
\cline{2-3}
&  Multi Finding \newline Grounding & Identifies the presence of candidate findings and provides the bounding box coordinates for each detected finding.\\
\cline{2-3}
&  Organ Grounding & Provides the bounding box coordinates for a given organ.\\
\cline{2-3}
&  Grounded Organ & Identifies the organ corresponding to a given bounding box. \\
\cline{2-3}
&  Grounded Phrase \newline Generation & Generates a radiology report phrase for the region corresponding to a given bounding box. \\
\cline{2-3}
&  Phrase Grounding & Provides the bounding box coordinates for the region described by a given phrase.\\
\cline{2-3}
&  Anatomical Region \newline Grounding & Provides the bounding box coordinates for a given anatomical region. \\
\cline{2-3}
& Grounded \newline Anatomical Region & Identifies the anatomical region corresponding to a given bounding box. \\
\midrule
\multirow{5}{=}{Visual Question Answering} & Visual Question \newline Answering & Responds to various questions about the content of a radiology image.\\
\cline{2-3}
&  Difference Visual \newline Question Answering & Compares past (reference) and current (main) images to answer questions about the differences between them. \\
\cline{2-3}
&  Visual \newline Instruction-following & Generates responses by following specific instructions, facilitating clear communication in multiple interactions. \\
\bottomrule
\end{NiceTabular}
}
\caption{Task Description.}
\label{tab:dataset-description}
\end{table*}

\begin{table*}[ht]
\renewcommand{\arraystretch}{1.0}
\centering
\scalebox{0.85}{
\begin{NiceTabular}{p{0.23\textwidth}|p{0.77\textwidth}}
\toprule
Task & Conversation Template \\ 
\midrule
\midrule
\textit{Single-image} & User: radiology image: \textbf{\textless image\textgreater} \ Which of the following findings are present in the radiology images? Findings: \{\textbf{findings}\} \newline 
Assistant: \{\textcolor{red}{\textbf{findings}}\}\newline
User: Based on the previous conversation provide a description of the findings in the radiology image. \newline
Assistant: \{\textcolor{red}{\textbf{report}}\}\\
\midrule
\textit{Multi-image} & User: radiology images: \{\textbf{images}\} Which of the following findings are present in the radiology images? Findings: \{\textbf{findings}\}
\newline 
Assistant: \{\textcolor{red}{\textbf{findings}}\}\newline
User: Based on the previous conversation provide a description of the findings in the current follow-up radiology images. \newline
Assistant: \{\textcolor{red}{\textbf{report}}\}\\
\midrule
\textit{Multi-study} & User: prior radiology images: \{\textbf{prior images}\} prior radiology report: \{\textbf{prior report}\} \newline 
follow-up images: \{\textbf{follow-up images}\} The radiology studies are given in chronological order. Which of the following findings are present in the current follow-up radiology images? Findings: \{\textbf{findings}\}\newline 
Assistant: \{\textcolor{red}{\textbf{findings}}\}\newline
User: Based on the previous conversation provide a description of the findings in the current follow-up radiology images. \newline
Assistant: \{\textcolor{red}{\textbf{report}}\}\\
\midrule
Disease Classification & User: radiology image: \textbf{\textless image\textgreater} Which of the following findings are present in the radiology image? Findings: \{\textbf{findings}\} \newline Assistant: \{\textbf{\textcolor{red}{findings}}\} \\
\midrule
Finding Grounding & User: radiology image: \textbf{\textless image\textgreater} Is \{\textbf{finding}\} present in the radiology image? If so, provide the bounding box coordinates of the region. \newline Assistant: \{\textbf{\textcolor{red}{bbox}}\} \\
\midrule
Grounded Finding & User: radiology image: \textbf{\textless image\textgreater} Provide a finding name for this region. \{\textbf{bbox}\} \newline Assistant:\{\textbf{\textcolor{red}{finding}}\} \\
\midrule
Abnormality Detection & User: radiology image: \textbf{\textless image\textgreater} Provide the bounding box coordinates of abnormal regions in the radiology image. \newline Assistant: \{\textbf{\textcolor{red}{bbox}}\} \\
\midrule
Multi Finding Grounding & User: radiology image: \textbf{\textless image\textgreater} Which of the following findings are present in the radiology image? Provide the bounding box coordinates if present. Findings: \{\textbf{findings}\} \newline
Assistant: \{\textbf{\textcolor{red}{findings, bboxes}}\} \\
\midrule
Organ Grounding & User: radiology image: \textbf{\textless image\textgreater} Provide the bounding box coordinates of \{\textbf{organ}\} in the radiology image. \newline Assistant: \{\textbf{\textcolor{red}{bbox}}\} \\
\midrule
Grounded Organ & User: radiology image: \textbf{\textless image\textgreater} Provide an organ name for this region. \{\textbf{bbox}\} \newline Assistant: \{\textbf{\textcolor{red}{organ}}\} \\
\midrule
Grounded Phrase Generation & User: radiology image: \textbf{\textless image\textgreater} Provide a radiology report phrase for the region. \{\textbf{bbox}\} \newline Assistant: \{\textbf{\textcolor{red}{phrase}}\} \\
\midrule
Phrase Grounding & User: radiology image: \textbf{\textless image\textgreater} Provide the bounding box coordinate of the region this phrase describes: \{\textbf{phrase}\} \newline Assistant: \{\textbf{\textcolor{red}{bbox}}\} \\
\midrule
Anatomical Region \newline Grounding & User: radiology image: \textbf{\textless image\textgreater} Provide the bounding box coordinate of the anatomical region. \{\textbf{name}\} \newline Assistant: \{\textbf{\textcolor{red}{bbox}}\} \\
\midrule
Grounded \newline Anatomical Region & User: radiology image: \textbf{\textless image\textgreater} Provide an anatomical region name for this region. \{\textbf{bbox}\} \newline Assistant: \{\textbf{\textcolor{red}{name}}\} \\
\midrule
Visual Question Answering & User: radiology image: \textbf{\textless image\textgreater} Answer the question. \{\textbf{question}\} \newline Assistant: \{\textbf{\textcolor{red}{answer}}\} \\
\midrule
Difference Visual \newline Question Answering & User: reference: \textbf{\textless image\textgreater} main: \textbf{\textless image\textgreater} Using the provided reference and main radiology images answer the following question. \{\textbf{question}\} \newline Assistant: \{\textbf{\textcolor{red}{answer}}\} \\
\midrule
Visual Instruction-following & User: radiology image: \textbf{\textless image\textgreater} 
\{\textbf{question}\} 
\newline Assistant: \{\textbf{\textcolor{red}{answer}}\} \\
\bottomrule
\end{NiceTabular}
}
\caption{Conversation template for each task. 
Phrases inside brackets \{\} are replaced with appropriate text based on data instances, with red-colored text indicating target outputs.
\textless \textbf{image}\textgreater{} indicates the positions where visual tokens are inserted.
}

\label{tab:dataset-tem}
\end{table*}

\begin{table*}[ht]
\centering
\scalebox{0.9}{
\begin{NiceTabular}{p{0.12\textwidth}|p{0.2\textwidth}|p{0.2\textwidth}|p{0.1\textwidth}|p{0.13\textwidth}|p{0.13\textwidth}}
\toprule
Task type & Task & Dataset & Train & Dataset ratio & Task type ratio\\ 
\midrule
\midrule
\multirow{3}{=}{Medical Report Generation} & \textit{Single-image} & MIMIC-CXR & 270,236 & 200.0 &  \\ \cline{2-5}
                                           & \textit{Multi-image} & MIMIC-CXR & 151,606 & 120.0 & 0.54 \\ \cline{2-5}
                                           & \textit{Multi-study} & MIMIC-CXR & 68,373 & 60.0 &  \\
                                           
\midrule
\multirow{44}{=}{Image Understanding} & \multirow{7}{=}{Disease Classification} 
                                      & BRAX & 40,965 & 7.0 &  \\
                                      & & CheXpert & 223,414  & 30.0 & \\
                                      & & VinDr-CXR & 15,000 & 3.0 & \\
                                      & & ChestX-ray14 & 86,523  & 10.0 & \\
                                      & & ChestX-Det10 & 3,578  & 1.0 &  \\
                                      & & SIIM & 10,675 & 4.0 &  \\
                                      & & RSNA & 14,863 & 4.0 &  \\
                                      & & COVID-19 Radiography & 15,153  & 3.0 &  \\
\cline{2-5}
                                      & \multirow{8}{=}{Finding Grounding} & VinDr-CXR & 15,000 & 2.0 &  \\
                                      & & ChestX-ray14 & 50,500 & 10.0 &  \\
                                      & & ChestX-Det10 & 3,578 & 2.0 &  \\
                                      & & JSRT & 247 & 0.1 &  \\
                                      & & SIIM & 10,675  & 4.0 &  \\
                                      & & RSNA & 14,863 & 1.0 &  \\
                                      & & COVID-QU-Ex & 2,796  & 1.0 &  \\
                                      & & QaTa-COV19 & 4,194  & 1.0 &  \\
\cline{2-5}
                                      & \multirow{7}{=}{Grounded Finding} & VinDr-CXR & 4,394  & 2.0 &  \\
                                      & & ChestX-Det10 & 2,967 & 2.0 &  \\
                                      & & JSRT & 154 & 0.1 &  \\
                                      & & SIIM & 2,379 & 2.0 &  \\
                                      & & RSNA & 6,012  & 1.0 &  0.35\\
                                      & & COVID-QU-Ex & 1,864  & 1.0 &  \\
                                      & & QaTa-COV19 & 4,194 & 3.0 &  \\
\cline{2-5}
                                      & \multirow{7}{=}{Abnormality Detection} & VinDr-CXR & 4,394 & 3.0 &  \\
                                      & & ChestX-Det10 & 2,967  & 1.5 &  \\
                                      & & JSRT & 154 & 0.1 &  \\
                                      & & SIIM & 2,379  & 2.0 &  \\
                                      & & RSNA & 6,012  & 1.0 &  \\
                                      & & COVID-QU-Ex & 1,864  & 1.5 &  \\
                                      & & QaTa-COV19 & 4,194  & 1.5 &  \\
\cline{2-5}
                                      & \multirow{3}{=}{Multi Finding Grounding} & ChestX-ray14 & 50,500  & 10.0 &  \\
                                      & & VinDr-CXR & 15,000  & 2.0 &  \\
                                      & & ChestX-Det10 & 3,578 & 2.0 &  \\
\cline{2-5}
                                      & \multirow{2}{=}{Organ Grounding} & COVID-19 Radiography & 15,153  & 0.8 &  \\
                                      & & COVID-QU-Ex & 3,728  & 0.8 &  \\
\cline{2-5}
                                      & \multirow{2}{=}{Grounded Organ} & COVID-19 Radiography & 15,153 & 0.8 &  \\
                                      & & COVID-QU-Ex & 3,728 & 0.8 &  \\
\cline{2-5}
                                      & \multirow{2}{=}{Grounded Phrase Generation} & MS-CXR & 638  & 2.0 &  \\
                                      & & ImaGenome & 164,229 & 40.0 &  \\
\cline{2-5}
                                      & \multirow{2}{=}{Phrase Grounding} & MS-CXR & 638 & 2.0 &  \\
                                      & & ImaGenome & 164,229 & 40.0 &  \\
\cline{2-5}
                                      & Anatomical Region & \multirow{2}{=}{ImaGenome} & \multirow{2}{=}{164,229} & \multirow{2}{=}{20.0} &  \\
                                      & Grounding & & & &  \\
\cline{2-5}
                                      & Grounded & \multirow{2}{=}{ImaGenome} & \multirow{2}{=}{164,229} & \multirow{2}{=}{20.0} &  \\
                                      & Anatomical Region & & & &  \\
\midrule
\multirow{6}{=}{Visual Question Answering} & \multirow{2}{=}{Visual Question Answering} & MIMIC-CXR-VQA & 255,919 & 2.0 &  \\
                                           & & MIMIC-Diff-VQA & 553,156  & 2.0 &  \\
\cline{2-5}
                                      & Difference Visual & \multirow{2}{=}{MIMIC-Diff-VQA} & \multirow{2}{=}{129,900} & \multirow{2}{=}{4.0} & \multirow{2}{=}{0.11} \\
                                      & Question Answering & & & &  \\
\cline{2-5}
                                      & Visual & \multirow{2}{=}{RaDialog} & \multirow{2}{=}{297,964} & \multirow{2}{=}{70.0} & \\
                                      & Instruction-following & & & &  \\
\bottomrule
\end{NiceTabular}
}
\caption{
The number of training instances and 
sampling ratios
for each dataset used
across tasks.
}
\label{tab:ratio}
\end{table*}

\end{document}